\documentclass{article}

\usepackage[margin=30truemm]{geometry}
\usepackage{microtype}
\usepackage{graphicx}
\usepackage{subfigure}
\usepackage{booktabs} 
\usepackage{hyperref}
\usepackage{algorithm}
\usepackage{algorithmic}

\usepackage{color}
\usepackage{lipsum}

\newcommand{\argmin}{\mathop{\rm arg~min}\limits}

\DeclareFontShape{OT1}{cmr}{bx}{sc}{<-> cmbcsc10}{}

\title{Optimal Layer Selection for Latent Data Augmentation}
\author{Tomoumi Takase, Ryo Karakida}

\begin{document}

\maketitle

\begin{abstract}
While data augmentation (DA) is generally applied to input data, several studies have reported that applying DA to hidden layers in neural networks, i.e., feature augmentation, can improve performance. However, in previous studies, the layers to which DA is applied have not been carefully considered, often being applied randomly and uniformly or only to a specific layer, leaving room for arbitrariness. Thus, in this study, we investigated the trends of suitable layers for applying DA in various experimental configurations, e.g., training from scratch, transfer learning, various dataset settings, and different models. In addition, to adjust the suitable layers for DA automatically, we propose the adaptive layer selection (AdaLASE) method, which updates the ratio to perform DA for each layer based on the gradient descent method during training. The experimental results obtained on several image classification datasets indicate that the proposed AdaLASE method altered the ratio as expected and achieved high overall test accuracy.
\end{abstract}

\section{Introduction}
Data augmentation (DA) techniques are indispensable for deep-learning tasks, and many DA methods have been developed \cite{yun_cutmix_2019,bao_dp-mix_2023,hou_when_2023}. In several basic and applied studies, DA improved generalization performance of models. However, the inappropriate use of DA techniques deteriorates it, as seen in the study of \cite{yun_cutmix_2019}. An effective way to use DA should be reconsidered.

One issue to consider is the heuristic selection of layers where the DA is applied. While DA is applied in an input space in most cases, some DA methods augment samples in latent spaces, e.g., the manifold mixup method \cite{verma_manifold_2019}. The manifold mixup method applies mixup \cite{zhang_mixup_2018}, i.e., the linear interpolation of two samples, in layers selected randomly in a neural network. However, the manifold mixup algorithm does not consider suitable layers to apply data augmentation, and the corresponding study did not address the layer suitability problem for DA, which makes layer selection for DA highly heuristic. Thus, it remains uncertain whether a randomly selected layer or a specific layer is more appropriate for applying mixup. Furthermore, exploring whether the improvement of the training performance by latent-DA is limited to mixup or extends to other DA methods such as masking and affine transformation is an interesting issue to advance DA. Feature augmentation, which involves applying Gaussian noise to a specific layer such as the last layer, has been proposed in a previous study \cite{li_simple_2021}, but it also encounters the same issue. 

In this work, we mainly made the following three contributions. First, we proposed a method that realizes automatic search for suitable layers. Searching those layers automatically is expected to reduce the computational costs of finding suitable layers. For this task, we propose the adaptive layer selection (AdaLASE) method, which searches appropriate layers for DA dynamically during training. The proposed method adjusts the acceptance ratio defined for each layer, which is used to determine the layer for DA probabilistically, using the gradient descent method to minimize the validation loss. In practice, the losses for the augmented training data are used instead of the validation loss because these losses tend to have a high correlation. The detail of the proposed AdaLASE method is described in Section 4. The final AdaLASE equation is given by Eq. (7), and it is also possible to implement the AdaLASE code according to Algorithm 1.

Second, we investigated the performance of the proposed AdaLASE method compared with applying DA to the input data or randomly selected layers and whether AdaLASE works effectively by plotting the transitions of the acceptance ratio during training. The problem of selecting suitable layers is complex because training performance largely depends on the experimental settings; however, we can conclude that the proposed AdaLASE method is a promising solution for searching layers in a dynamic manner. Then, experimental results are reported in Sections 5.1, 5.2, and 5.3. 

In the experiments, we focused on whether AdaLASE is able to select a suitable acceptance ratio. Since conventional methods may have suitable acceptance ratios, AdaLASE does not necessarily need to achieve better accuracy than conventional methods; it is important that it achieves high accuracy overall, regardless of conditions. The results in Table 1 show that AdaLASE performed as well as or better than Uniform DA. The results in Figures 3 and 5 also show that AdaLASE was able to increase an acceptance ratio for layers suitable for data augmentation during training.

Lastly, we investigated the trends of suitable layers for different experimental configurations through comprehensive experiments, considering that suitable layers for DA will differ depending on the experimental settings, e.g., the datasets and models used in the experiments. This leads to the elimination of heuristic layer selection in DA, thereby reducing the computational costs associated with searching layers and improving training performance.Then, experimental results are reported in Section 5.4. 

In experiments described in Section 5.4, pretrained models were used because models are frequently trained using transfer learning. In the experiments with transfer learning, suitable layers for DA were investigated during a fine-tuning process. In addition, we focused on a small number of training samples in those experiments because sufficient numbers of samples are frequently difficult to obtain, and effective training with small numbers of samples is a critical challenge. In such situations, the application of DA effectively contributes considerably to the realization of high performance. Although a small number of samples can be used for training from scratch, such samples are frequently used in transfer learning; thus, we investigated the effect of DA on a small number of samples using transfer learning. We performed the trainings using data augmentation with various acceptance ratios shown in Figure 6. The results in Table 2 and Figure 7 show that the effect of the data augmentation layer is highly dependent on the number of training samples.

\section{Related Work}
In addition to manifold mixup \cite{verma_manifold_2019}, several previous studies have addressed DA in feature spaces. For example, DeVries et al.\cite{devries_dataset_2017} applied Gaussian noise, interpolation, or extrapolation to feature vectors obtained using a long short-term memory autoencoder. In a study by Chu et al. \cite{chu_feature_2020}, samples for head classes and tail classes were mixed in feature maps. Note that the methods used in these studies require models with specific structures, e.g., autoencoders, and specific datasets, e.g., imbalanced data. The Shake-shake \cite{gastaldi_shake-shake_2017} method applies DA in a hidden space by modifying the structure of residual networks, whereas the ShakeDrop method \cite{yamada_shakedrop_2019} combines Shake-shake and dropout to augment samples in hidden spaces. However, these methods require the use of residual networks. In contrast, the current study did not limit the types of data, DA methods, or neural network structures.

Previous studies have proposed latent-DA methods that are not strongly dependent on the structure of the model. For example, in DropBlock study \cite{ghiasi_dropblock_2018}, dropout was performed in latent spaces. In addition, in CutMix \cite{yun_cutmix_2019}, which combines cutout \cite{devries_improved_2017} and mixup, CutMix was performed in latent spaces. Li et al. \cite{li_simple_2021} proposed a simple technique that perturbs the feature embedding with Gaussian noise during training. As demonstrated by these studies, some have investigated DA for latent spaces during training with general neural networks. In the current study, we demonstrated that applying DA to hidden layers improves the training performance regardless of the DA methods, and we attempted to identify the common characteristics of suitable layers for DA in several experimental configurations.

Several studies have investigated ways to identify appropriate DA settings automatically. For example, the AutoAugment method \cite{cubuk_autoaugment_2019}, which was inspired by advances in neural architecture search methods \cite{liu_progressive_2018,pham_efficient_2018,zoph_neural_2022}, finds appropriate DA policies using reinforcement learning and optimizes the parameters of each DA method automatically. Note that AutoAugment is accurate; however, it is also a time-intensive method. The Fast AutoAugment method proposed by Lim et al. \cite{lim_fast_2019} accelerates the parameter search of the ImageNet dataset from 15,000 GPU hours with AutoAugment to 450 GPU hours; however, its computational time is considerably longer than that of conventional augmentation techniques. Thus, Fast AutoAugment cannot replace conventional augmentation techniques, which are easy to use without incurring excessive computational costs.

RandAugment \cite{cubuk_randaugment_2020} is also an application of AutoAugment that selects some augmentations randomly from 14 distinct augmentations. Here, the number of manipulations used simultaneously and the magnitude of deformation for the DA must be determined before training, and these augmentation settings are determined for each minibatch. Compared with AutoAugment, the RandAugment method reduces computational costs considerably. The UniformAugment \cite{lingchen_uniformaugment_2020} and TrivialAugment \cite{muller_trivialaugment_2021} methods are improved versions of the RandAugment method. In addition, AutoDropout \cite{pham_autodropout_2021} automates the process of designing dropout patterns. Note that RandAugment has a simple algorithm; however, it has a disadvantage because it is used heuristically.

As a method to optimize DA policies during training, we previously proposed the self-paced augmentation method \cite{takase_self-paced_2021}, which selects samples to apply DA and was inspired by curriculum learning and self-paced learning. The OnlineAugment method \cite{vedaldi_onlineaugment_2020} optimizes three complementary augmentation models responsible for different types of augmentations, and the Meta Approach to Data Augmentation Optimization (MADAO) method \cite{hataya_meta_2022} is a dynamic augmentation method that optimizes DA policies using an implicit gradient with Neumann series approximation. The proposed AdaLASE method differs from these existing methods because it selects suitable layers for DA applications.

\section{Latent Data Augmentation}
In a convolutional neural network (CNN), features are extracted hierarchically. For example, a CNN has a nonlinear structure (including convolution layers, pooling layers, and activation function); thus, applying DA to different layers can produce different feature maps with the same labels. Therefore, applying DA to the hidden layers will yield an effect that differs from applying DA to the input layer. To apply DA to the hidden layers, appropriate candidate positions for applying DA must be determined.

\begin{figure}
\hfill
\begin{center}
\subfigure[MLP]{
\includegraphics[width=1.0in]{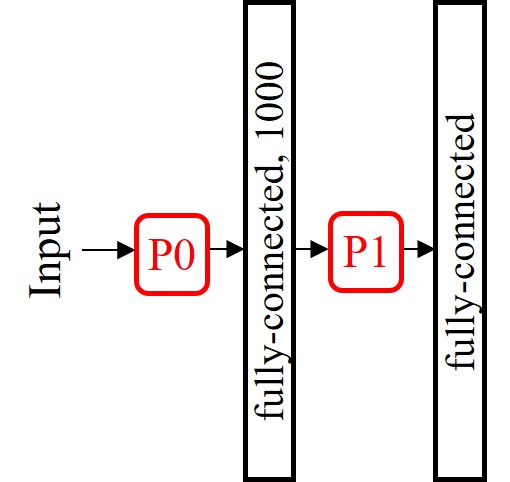}
}
\subfigure[ResNet]{
\includegraphics[width=3.5in]{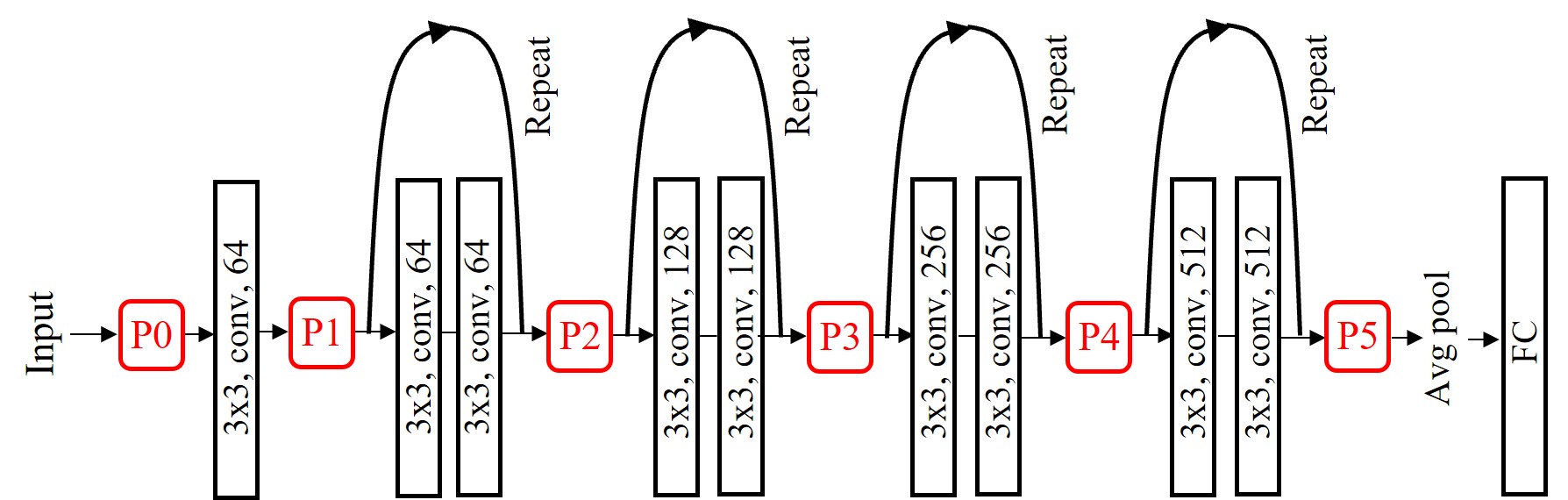}
}\\
\end{center}
\caption{Positions to apply DA in several neural networks.}
\label{fig:resnet}
\end{figure}

We selected the position for DA as P$i$ (where $i$ is 0 $\cdots$ $K$). The positions to apply DA in several neural networks are shown in Fig. 1. For example, for ResNet, we inserted P$i$ into the positions outside a residual block, although DA processes can be inserted at any position.

Latent-DA is performed as follows: First, data are input and forwardly propagated until the $n$th hidden layer, which is determined in advance. Then, DA is performed at the hidden layer. Here, the DA parameters, e.g., the position and size of the mask in random erasing, should be common to all channels. In this case, the deformation in the feature maps that emerges in the subsequent convolution process is similar to that induced by DA. If the DA parameters differ in each channel, the samples are deformed excessively, the training performance deteriorates, and augmentation of the same image in several layers is inappropriate for the same reason. Finally, the augmented data are forwardly propagated until the output layer.

This method can be implemented easily and does not increase training time significantly. DA can also be applied in layers that are selected randomly for each minibatch, e.g., a manifold mixup. We refer to this technique as uniform augmentation because the layer is selected based on the uniform distribution. This functionality is expected to increase the distribution range of the training data and improve the model's generalizability. However, appropriate layers for DA differ for different datasets, and applying DA at inappropriate layers should be avoided. Thus, one goal of the current study is to identify the trends of the experimental configurations for which uniform augmentation works effectively.

\begin{figure}
\hfill
\begin{center}
\includegraphics[width=4.5in]{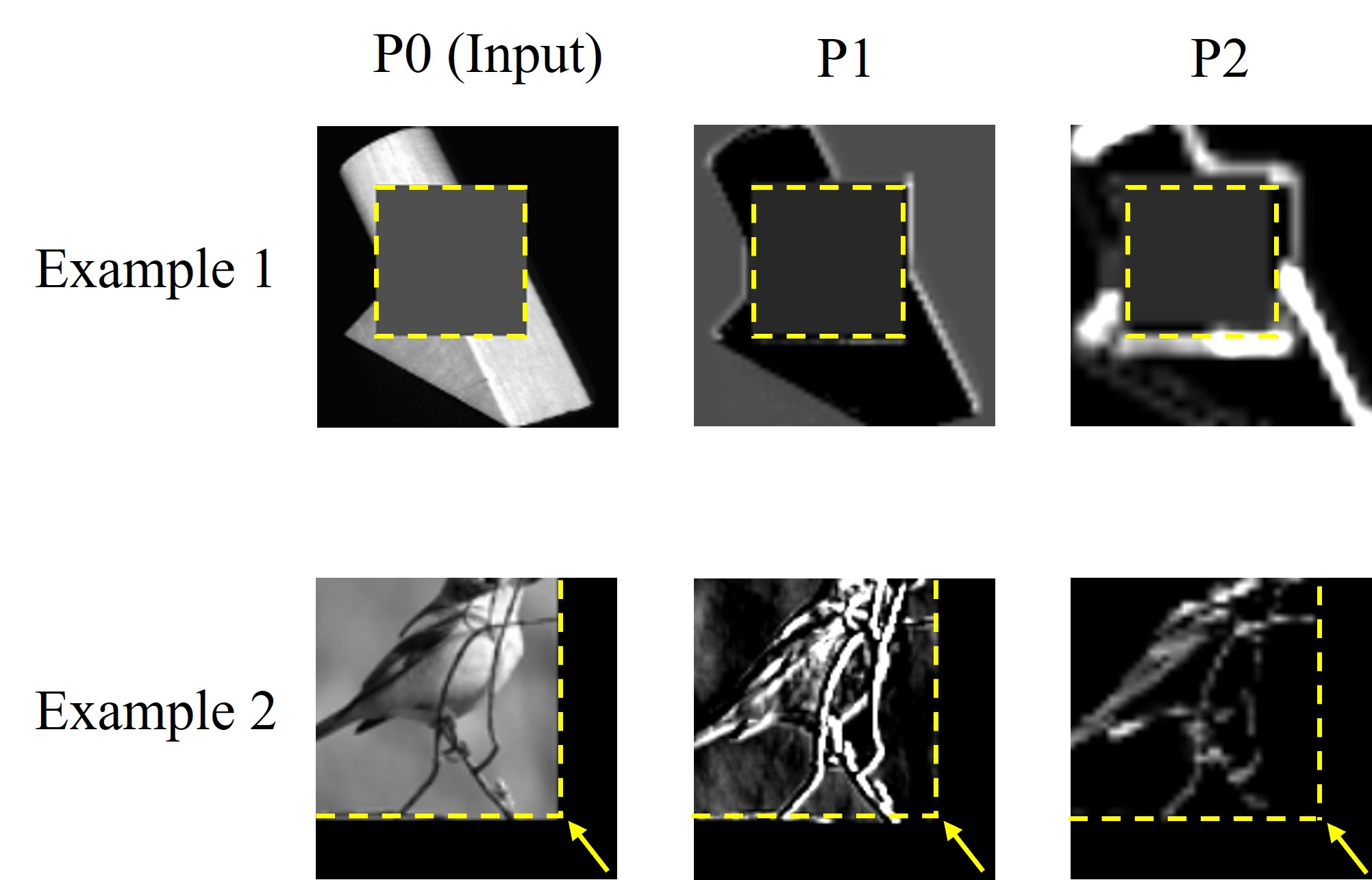}
\end{center}
\caption{Examples of latent-DA. The sample in Example 1 was selected from the COIL-20 dataset, and the sample in Example 2 was selected from the STL-10 dataset. These samples were augmented using cutout or translation at P0, P1, and P2 in ResNet18.}
\label{fig:examples}
\end{figure}

Fig. 2 shows the actual feature maps obtained by applying DA on the COIL-20 and STL-10 datasets. Here, cutout was applied to Example 1, and translation was applied to Example 2. P0 denotes the input layer, and P1 and P2 denote the hidden layers shown in Fig. 1(b). We can apply DA to the feature maps obtained in P1 or P2, just like applying that to P0. Note that latent-DA is not limited to image data.

Transfer learning is frequently used in many practical applications; thus, investigating the layer to apply DA is significant for fine-tuning processes. Features are obtained in the pretraining process of transfer learning; thus, destroying those features by applying DA reduces performance.

\section{Proposed AdaLASE Method}
Although latent augmentation for general DA methods (Section 3) is one of our proposed methods, we improved the latent augmentation so that DA is applied intensively in suitable layers. To determine suitable layers for DA automatically and dynamically, we also propose the AdaLASE method. In the proposed method, we define the acceptance ratio for layer $i$ as $q_i$ such that $\sum_{i,q_i} =1$ $(0 \leq q_i \leq 1)$. Here, the acceptance ratio is updated for each iteration, and the layer for DA is determined stochastically based on the acceptance ratio.

The training loss using $q_l$ is expressed as follows:

\begin{eqnarray}
\label{eq:AdaLASE1}
L_{train} = \sum_{i=1}^K q_i L_{train, i},
\end{eqnarray}

\noindent where $q_i$ is set to the same value for all layers in the initial setting and is updated as training proceeds.

When layer $l$ is selected for DA, we update the acceptance ratio $q_l$ as follows using a gradient descent--based method:

\begin{eqnarray}
\label{eq:AdaLASE2}
q_l \leftarrow q_l - \eta \frac{\partial L_{val}}{\partial q_l},
\end{eqnarray}

\noindent where $L_{val}$ is the validation loss, and $\eta$ is the step size. Reducing the validation loss, i.e., the value of the loss function when using validation data, is a purpose of training; thus, modifying $q_l$ such that the loss is reduced is a reasonable approach. Here, $\partial L_{val} / \partial q_l$ is deformed using $\theta^{\ast}$, which is defined by $\theta^{\ast} = \argmin_{\theta} L_{train}(\theta, q_l)$, as follows:

\begin{eqnarray}
\label{eq:AdaLASE3}
\frac{\partial L_{val}}{\partial q_l} = \frac{\partial L_{val}}{\partial \theta^{\ast}} \frac{\partial \theta^{\ast}}{\partial q_l}.
\end{eqnarray}

\noindent Here, to calculate $\partial \theta^{\ast} / \partial q_l$, we consider the following equation:

\begin{eqnarray}
\label{eq:AdaLASE4}
\frac{\partial}{\partial \theta^{\ast}} L_{train}(\theta^{\ast}, q_l) = 0.
\end{eqnarray}

\noindent By defining the left side of this equation as $f$, this can be deformed using the implicit function theorem as follows:

\begin{eqnarray}
\label{eq:AdaLASE5}
\frac{\partial \theta^{\ast}}{\partial q_l} = -\left[\frac{\partial^2 L_{train}}{\partial \theta^{\ast} \partial \theta^{\ast}}\right]^{-1} \frac{\partial^2 L_{train}}{\partial \theta^{\ast} \partial q_l}.
\end{eqnarray}

\noindent Computing the inverse matrix of the Hessian matrix in deep learning is difficult because it contains a large number of parameters. In this case, we take the simplest approach, i.e., the Hessian matrix is approximated by the identity matrix. This approximation is referred to as the $T2-T1$ method \cite{luketina_scalable_nodate} in the hyperparameter optimization field and frequently works effectively despite its simplicity. To the best of our knowledge, the proposed method represents the first attempt to apply such gradient--based hyperparameter optimization to the layer selection task for latent-DA.

Although there are other sophisticated approximations, e.g., the Neumann series \cite{lorraine_optimizing_2020}, we employed the $T2-T1$ method because it possesses much better computational scalability. Then, Eq. \ref{eq:AdaLASE5} is deformed as follows:

\begin{eqnarray}
\label{eq:AdaLASE6}
\frac{\partial \theta^{\ast}}{\partial q_l} = -\frac{\partial}{\partial \theta^{\ast}}\left(\frac{\partial L_{train}}{\partial q_l}\right) = -\frac{\partial L_{train, l}}{\partial \theta^{\ast}},
\end{eqnarray}
where the last transformation is specific to our latent-DA method. The transformation is the advantage of our formulation of the objective function (1) and simplifies the computation of the gradient.

Note that $L_{val}$ in Eq. (\ref{eq:AdaLASE2}) requires the validation samples and preparing a sufficient number of such samples can be costly in actual training processes; thus, we propose using $L_{DA}$, which is the training loss for a dataset where DA has been applied to the input data. We refer to this alternative dataset as the pseudo-validation dataset, and training loss with DA correlates with test loss. Ultimately, Eq. (\ref{eq:AdaLASE2}) can be deformed using Eqs. (\ref{eq:AdaLASE1}) and (\ref{eq:AdaLASE6}) as follows:

\begin{eqnarray}
\label{eq:AdaLASE7}
q_l \leftarrow q_l + \eta \frac{\partial L_{DA}}{\partial \theta}^T \frac{\partial L_{train, l}}{\partial \theta}.
\end{eqnarray}
\noindent 
Note that computing $\theta^{\ast}$ at each step is highly demanding.
Therefore, we replace $\theta^{\ast}$ with  $\theta$ at each epoch for convenience as is usual in gradient-based hyper-parameter optimization. In summary, updating the acceptance ratio requires the partial differential of the training loss for a pseudo-validation dataset and that of the training loss with the target DA applied at layer $l$ based on the acceptance ratio. When using the proposed AdaLASE method, the update amount can be averaged over several iterations to reduce the impact of randomness. In addition, we set a lower limit for the acceptance ratio so the ratio can increase again when it becomes close to 0. The lower limit for the acceptance ratio in each layer is set to $t_i/M$, where $t_i$ is a predetermined hyperparameter and $M$ is the total number of positions to apply DA. Note that we set $t_i$ to 0.1 in most of the experiments conducted in this study.

\begin{algorithm}[tb]                    
    \caption{Adaptive layer selection (AdaLASE)}
    \label{alg:adalase}
\begin{algorithmic}
    \STATE $L_{DA} \gets f(x_{DA, input}, \theta)$
    \STATE $l \gets {\rm Random}(q)$
    \STATE $L_{train, l} \gets f(x_{DA', l}, \theta)$
    \STATE $\theta_{t+1} \gets \theta_t - \lambda_t \nabla_\theta L_{train, l}(\theta_t)$
    \STATE $q_l \leftarrow q_l + \eta \frac{\partial L_{DA}}{\partial \theta}^T \frac{\partial L_{train, l}}{\partial \theta}$
	\IF {$q_l < d$}
	\STATE $q_l \gets d$
	\ENDIF
	\IF {$q_l > 1 - d$}
	\STATE $q_l \gets 1 - d$
	\ENDIF
	\FOR {$i$}
	\STATE $q_i \gets q_i / \sum{q}$
	\ENDFOR
	\STATE $\lambda_{t+1} \gets g(\lambda_t, t)$
\end{algorithmic}
\end{algorithm}

The algorithm utilized in the proposed AdaLASE method for each iteration is given in Algorithm 1, where $x_{DA, input}$ denotes a pseudo-validation dataset, $x_{DA', l}$ denotes data augmented in layer $l$, which is determined randomly based on each $q_i$, and $\lambda$ denotes the learning rate updated based on the function $g$. Here, $d$ is the lower limit for the acceptance ratio. If the acceptance ratio in a layer becomes 0, the layer will not be selected again; thus, the ratio cannot increase. To avoid this issue, we set the acceptance ratio for each layer such that it is not less than $d$. In most of the experiments conducted in this study, we used for $d = 0.1 / K$, where $K$ is the total number of DA positions.

The proposed AdaLASE method can also be used to optimize the types, e.g., cutout and mixup, and the hyperparameters, e.g., the mask size, of DA methods. This is easily realized by extending the algorithm used to adjust only the layers for DA. Training based on the suitability of DA methods for layers, types, or hyperparameters determined by the proposed AdaLASE method facilitates effective use of DA methods.

\section{Experiments}
\subsection{Accuracy Evaluation of Latent-DA}
Several experiments were conducted to verify the effect of latent-DA. In this section, we report the results of comprehensive experiments conducted to identify the trends of suitable layers for DA and experiments conducted to verify the algorithm employed in the proposed AdaLASE method. These experiments were performed to understand which layer is suitable for DA in each experimental configuration. Note that the purpose is not to realize state-of-the-art performance but understanding the characteristics of DA by investigating the trends of suitable layers in various experimental configurations. Thus, we employed simple experimental configurations, e.g., the models were trained with a single DA method, e.g., mixup or cutout. In addition, as well as AdaLASE, all latent-DA methods are the proposed methods, and we focused on finding practical uses rather than simply developing the proposed methods. Here, we compared the test accuracy when DA was applied to different layers on several datasets, models, DA methods, and numbers of samples. In training from scratch, the results for no DA, input DA, uniform DA, and the proposed AdaLASE method were compared and analyzed.

For DA methods, we used mixup \cite{zhang_mixup_2018} and cutout \cite{devries_improved_2017}. In mixup, the pixel values and labels of two samples were interpolated linearly based on the mix rate sampled from a beta distribution. Here, the hyperparameter $\alpha$ for the beta distribution was set to 1, when the beta distribution became the uniform distribution. Note that mixup is identical to manifold mixup when applied dynamically to feature maps in a randomly selected layer during training. In addition, cutout applied a randomly sized mask to part of the image, where the mask size was set to half the image size. In addition, random rotation was applied to generate $x_{DA, input}$, i.e., the pseudo-validation dataset in the proposed AdaLASE method. Here, the degree of rotation was selected randomly from [-10, 10]. Based on common settings, several DA methods were always applied to input data. Specifically, random crop and horizontal flip were used for the CIFAR-10, CIFAR-100, and SVHN datasets, and center crop, horizontal flip, and color tone change were employed for the ImageNet dataset.

As mentioned previously, the CIFAR-10, CIFAR-100, SVHN, and ImageNet datasets, which are frequently used to benchmark machine learning models, were used in these experiments. For models, we used the ResNet18 and ResNet50 CNN models. The models were trained over 200 epochs with a batch size of 256 for all datasets (except for ImageNet). For ImageNet, the model was trained over 100 epochs with a batch size of 2048. The test accuracy for each epoch was calculated using the original test data. In addition, the model parameters were updated using the momentum stochastic gradient descent method, and the learning rate was varied according to cosine annealing with an initial learning rate of 0.1. The initial weights were sampled from a uniform distribution. To moderate the effects of initialization on the results, we ran five trials with initial weights generated by different random seeds and calculated the mean and standard deviation for the best test accuracy.

\begin{table}[tb]
\caption{Test accuracy (\%) of the proposed AdaLASE and compared methods on several datasets. The highest accuracy in each training is shown. Mixup or cutout was used for DA, and the models were trained from scratch.}
\label{table:test_acc}
\vskip 0.15in
\begin{center}
\begin{small}
\begin{sc}
\begin{tabular}{lll}
\toprule
 & {\scriptsize CIFAR-10} & {\scriptsize CIFAR-100}\\
 & {\scriptsize ResNet18} & {\scriptsize ResNet18}\\
\hline
No DA              & 94.72 $\pm$ 0.25 & 75.67 $\pm$ 0.54\\
\hline
Input DA, Mixup    & 95.84 $\pm$ 0.23 & 78.01 $\pm$ 0.43\\
Uniform DA, Mixup  & 95.87 $\pm$ 0.09 & \textbf{79.86 $\pm$ 0.37}\\
AdaLASE, Mixup     & \textbf{95.96 $\pm$ 0.14} & 79.54 $\pm$ 0.34\\
\hline
Input DA, Cutout   & 95.59 $\pm$ 0.11 & 74.70 $\pm$ 0.47\\
Uniform DA, Cutout & 95.74 $\pm$ 0.22 & 76.04 $\pm$ 0.34\\
AdaLASE, Cutout    & \textbf{95.83 $\pm$ 0.18} & \textbf{76.19 $\pm$ 0.35}\\
\midrule
 & {\scriptsize SVHN} & {\scriptsize ImageNet}\\
 & {\scriptsize ResNet18} & {\scriptsize ResNet50}\\
\hline
No DA              & 96.47 $\pm$ 0.06 & 74.61 $\pm$ 0.38\\
\hline
Input DA, Mixup    & 96.90 $\pm$ 0.10 & 73.38 $\pm$ 0.58\\
Uniform DA, Mixup  & \textbf{97.24 $\pm$ 0.05} & 74.47 $\pm$ 0.14\\
AdaLASE, Mixup     & 97.23 $\pm$ 0.12 & \textbf{75.02 $\pm$ 0.19}\\
\hline
Input DA, Cutout   & 97.04 $\pm$ 0.12 & 75.09 $\pm$ 0.31\\
Uniform DA, Cutout & 97.11 $\pm$ 0.12 & 74.97 $\pm$ 0.41\\
AdaLASE, Cutout    & \textbf{97.16 $\pm$ 0.05} & \textbf{75.14 $\pm$ 0.41}\\
\bottomrule
\end{tabular}
\end{sc}
\end{small}
\end{center}
\vskip -0.1in
\end{table}

Table 1 shows the test accuracy obtained on several datasets. As can be seen, the uniform DA and proposed AdaLASE method outperformed the input DA method. Although the proposed AdaLASE obtained the highest accuracy in several cases, some of the experimental results confirmed that the accuracy of the AdaLASE method was worse than that of the uniform DA method. An important point is that AdaLASE achieved a performance that is close to the uniform DA rather the input DA. The proposed AdaLASE method selects an appropriate acceptance ratio dynamically; thus, the acceptance ratio does not always match the uniform distribution during training. This result implies that the acceptance ratio changed dynamically to decrease the validation loss does not necessarily improve the generalization performance of the model. However, even if the performance of the AdaLASE method is worse than that of uniform DA, it is not a problem if the difference is within the margin of error, as seen in these results. 

\subsection{Verification of AdaLASE Method based on Acceptance Ratio}
We also verified whether the proposed AdaLASE method works as expected by plotting the change in the acceptance ratio during training using the proposed method. If the acceptance ratio in AdaLASE changes to reduce the value of the loss function for the validation data, we consider that the proposed AdaLASE method worked as expected. To determine the magnitude of the validation loss, we computed the validation loss when applying DA to each layer during training. In this experiment, we trained MLP (Fig. 1(a)) on the CIFAR-10 dataset; thus, we computed the validation loss in two positions for DA, i.e., P0 and P1. During training with the proposed AdaLASE method, DA was applied at positions P0 and P1 in every epoch, and the validation loss was computed after updating the weights. The purpose of this experiment was to verify that the proposed AdaLASE method works effectively; thus, we used test data rather than a pseudo-validation dataset, which is required in AdaLASE.

\begin{figure}
\hfill
\begin{center}
\includegraphics[width=5.8in]{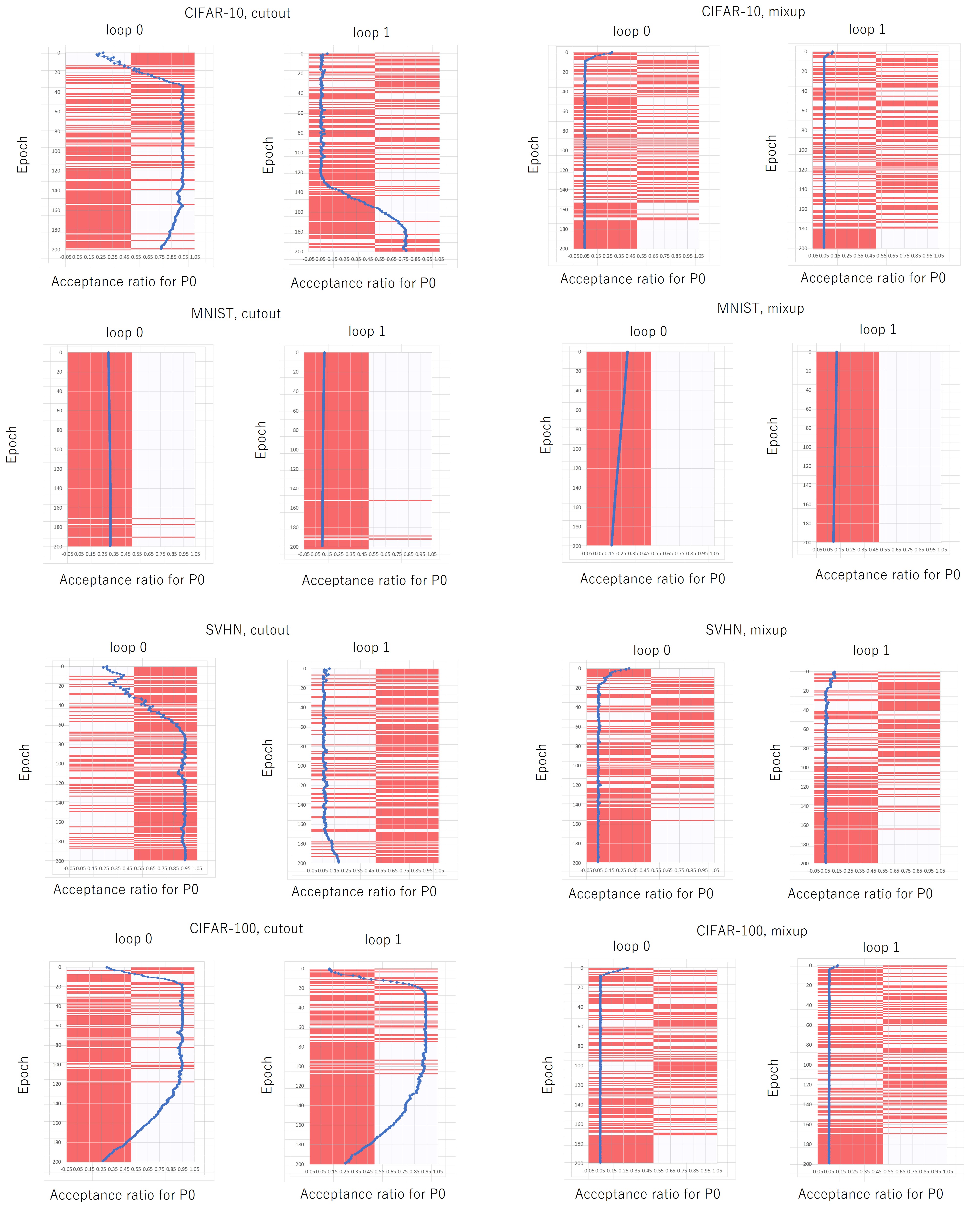}
\end{center}
\caption{Transitions of the acceptance ratio for P0 in proposed AdaLASE method when training MLP on the CIFAR-10 dataset. The results of 20 runs with different initializations are shown. Test data were used rather than the pseudo-validation data in the AdaLASE method. The models were trained from scratch.}
\label{fig:acceptance_ratio}
\end{figure}

The changes in the acceptance ratio while training using the proposed AdaLASE method are shown in Fig. 3, where the vertical axis represents the number of training epochs and training proceeds from top to bottom. Here, each dot represents the acceptance ratio for P0. Colors in the background were also based on the validation loss obtained during training with AdaLASE and represents which of P0 and P1 produced smaller validation loss when applying DA to each of P0 and P1 individually every epoch. The horizontal axis denotes the acceptance ratio of P0. Thus, if a dark color in the background is on the left, the validation loss for P1 is smaller than that for P0, and if the dark color is on the right, the validation loss for P0 is smaller than that for P1. In most results, the dots tended to change toward the dark color, i.e., the training processes were performed such that the layers that produced smaller validation losses were selected for DA. Thus, the proposed AdaLASE method was able to select suitable layers for DA, and these results demonstrate that the proposed method functioned properly.

\begin{figure}
\hfill
\begin{center}
\includegraphics[width=3.0in]{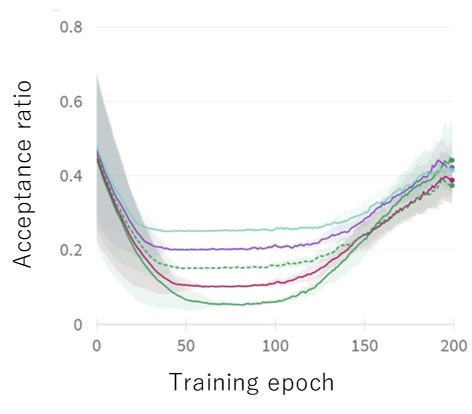}
\end{center}
\caption{Difference in acceptance ratios when the lower limit was varied. MLP was trained on the CIFAR-10 dataset with cutout. Test data were used rather than the pseudo-validation dataset in the AdaLASE method. The models were trained from scratch.}
\label{fig:lower_limit}
\end{figure}

We also investigated the effect of the lower limit of the acceptance ratio for each layer. Here, the acceptance ratios were compared when $Kd$ was set to 0.1, 0.2, 0.3, 0.4, or 0.5. In this evaluation, the MLP model was trained from scratch on the CIFAR-10 dataset, and cutout was applied during training using the proposed AdaLASE method.

\begin{figure}
\hfill
\begin{center}
\includegraphics[width=4.5in]{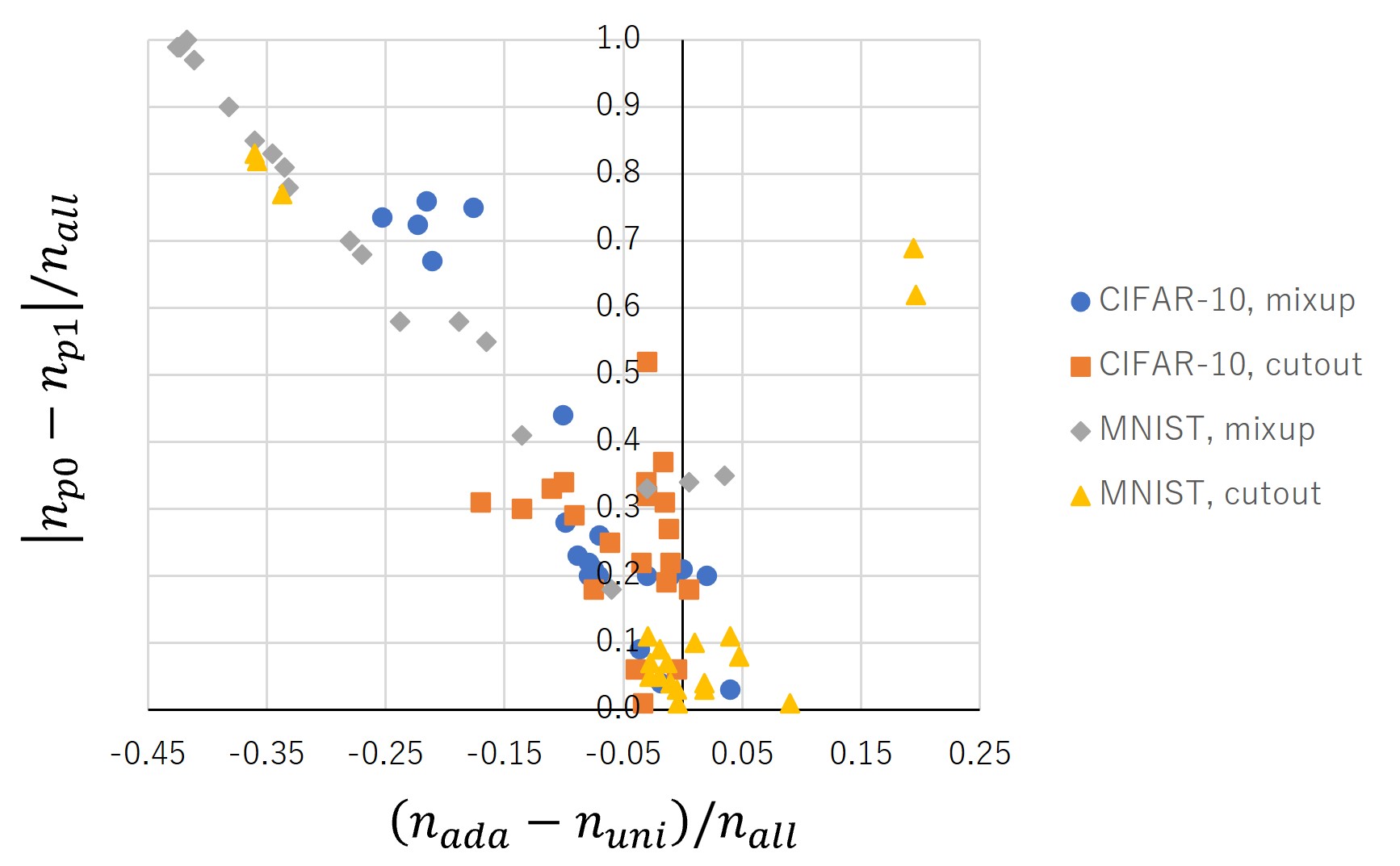}
\end{center}
\caption{Difference of numbers of the iterations that selected the worst layer between the AdaLASE method and the uniform ratio method. Test data were used rather than the pseudo-validation data in the AdaLASE method. The models were trained from scratch.}
\label{fig:number_worst}
\end{figure}

Fig. 4 shows the effect of the lower limit on the acceptance ratio while training using the proposed AdaLASE method. As can be seen, the difference in the lower limit did not affect the final ratio; thus, we do not need to search the lower limit strictly.

\subsection{Comparison of Layer Selection between AdaLASE and Uniform DA}
Although we confirmed that the proposed AdaLASE method tended to change the acceptance ratio successfully, we also quantitatively investigated whether the proposed AdaLASE method realized successful layer selection. Here, we counted the number of times that AdaLASE or the uniform DA selected the worst layer, i.e., the layer that produced the largest loss function values with the validation loss. Note that these results were obtained from the experiments discussed in the previous section; thus, the worst layer is either P0 or P1 in this experiment. Note that the MLP models were trained from scratch on the CIFAR-10 or MNIST datasets.

The corresponding results are shown in Fig. 5, where each point represents the result of each training process. Here, the horizontal axis is $(n_{ada}-n_{uni})/n_{all}$, where $n_{ada}$ denotes the number of times the worst layer was selected as the DA layer when using AdaLASE, $n_{uni}$ denotes the number of times the worst layer was selected as the DA layer when using uniform augmentation, and $n_{all}$ denotes the total number of iterations. The vertical axis is $|n_{p0}-n_{p1}|/n_{all}$, where $n_{p0}$ denotes the number of times that P0 was the worst layer, and $n_{p1}$ denotes the number of times that P1 was the worst layer. These results confirm that the proposed AdaLASE method selects the worst layer less frequently than the uniform augmentation method. In addition, the difference between $n_{p0}$ and $n_{p1}$ increased; thus, $(n_{ada}-n_{uni})$ tended to increase. These results further verify that the proposed AdaLASE method worked correctly.

\subsection{Sample Size Dependency in Transfer Learning}
\begin{figure}
\hfill
\begin{center}
\includegraphics[width=4.5in]{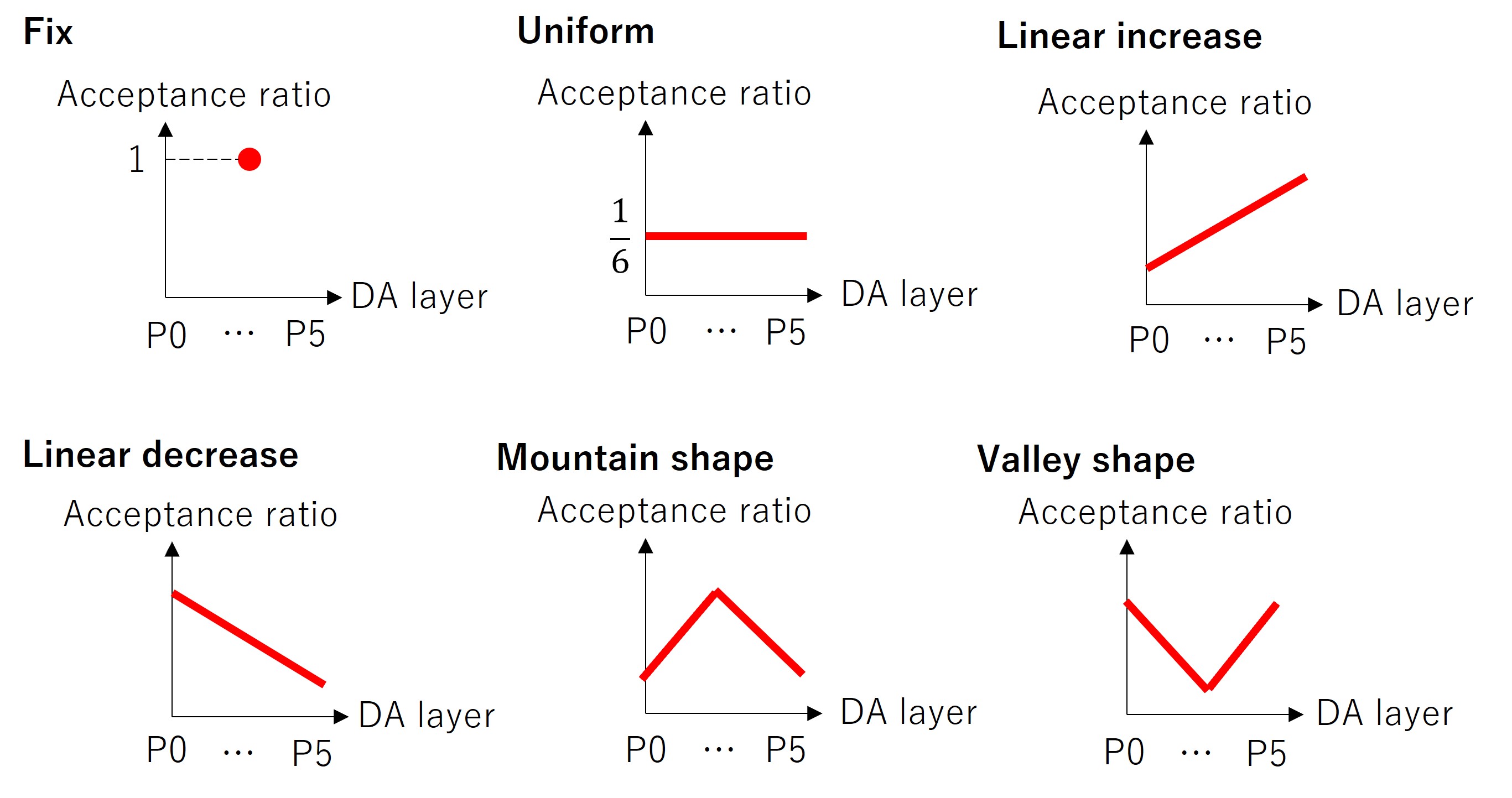}
\end{center}
\caption{Several methods to apply DA with different acceptance ratios.}
\label{fig:structure}
\end{figure}

In the experiments up to the previous section, we observed that AdaLASE tended to find optimal layers for DA. Thus, we investigated the layer suitability in detail on several experimental settings. In particular, we focused on sample size dependency in the transfer learning because the small number of samples is often used in transfer learning. The test accuracy of the ResNet18 model on the CIFAR-10 dataset was compared among various methods shown in Fig. 6. One method involved fixing the position to apply DA to P0 to P5 (Fig. 1(b)). In uniform augmentation, DA was applied randomly at a position selected from P0 to PM for a minibatch based on the uniform probability. In addition, DA was applied based on a linearly increasing or decreasing acceptance ratio, and the acceptance ratio was varied to become a mountain shape, which assigns a high ratio for layers that are close to the input and the output layers, or a valley shape, which assigns a low ratio for the same layers. In the proposed AdaLASE method, the position was determined based on the acceptance ratio for each position. For the hyperparameter settings of the proposed AdaLASE method, the initial acceptance ratio was set to 0.1, 0.01, and 0.001, and the step size of the mean for the ratio was set to 1 and 10. In these experiments, we investigated the effect of the number of training samples using 100, 1000, 10000, and 50000 (full) samples. Here, we used the pretrained ResNet18 model trained on the ImageNet dataset.

For DA methods, in addition to mixup \cite{zhang_mixup_2018} and cutout \cite{devries_improved_2017}, we used translation and CutMix \cite{yun_cutmix_2019}. Here, the translation method sampled two random values from [0, 0.2] and moved the image up, down, left, and right by a certain number of pixels (computed as the image size multiplied by the sampled values). The value of the hyperparameter $\alpha$ for mixup in the CutMix method was set to 0.5.

\begin{figure}
\hfill
\begin{center}
\subfigure[Mixup]{
\includegraphics[width=4.3in]{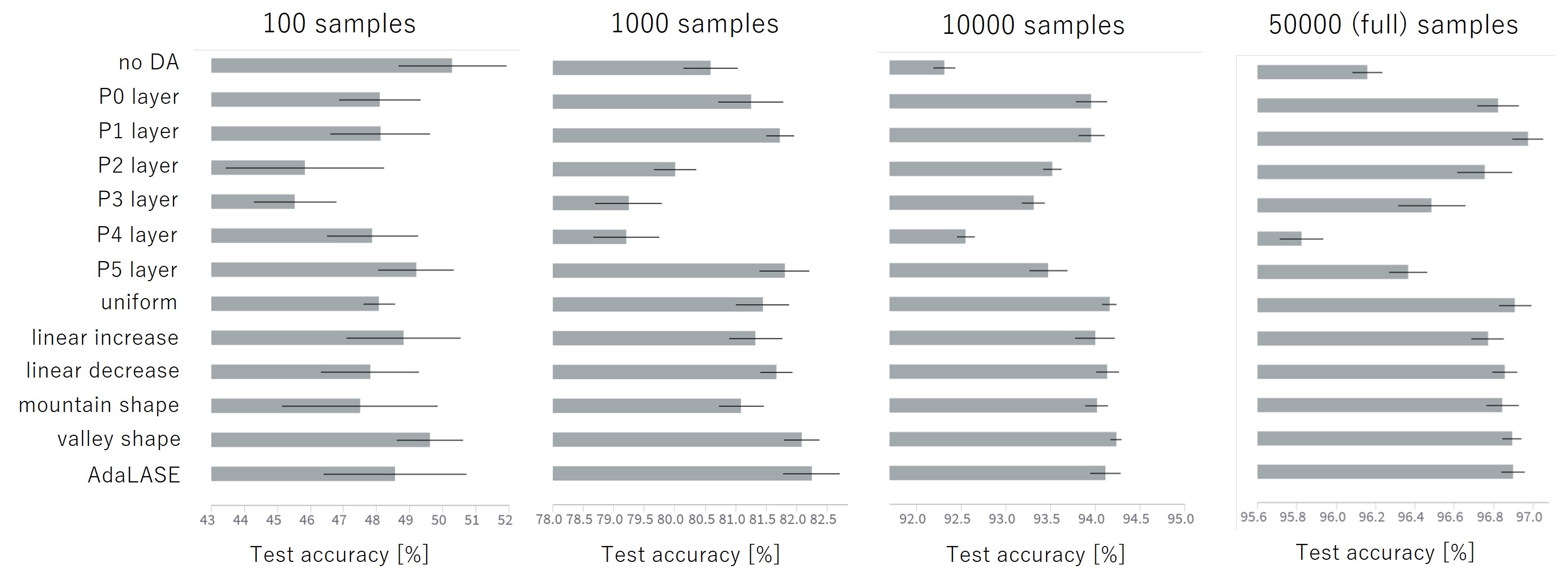}
}\\
\subfigure[Cutout]{
\includegraphics[width=4.3in]{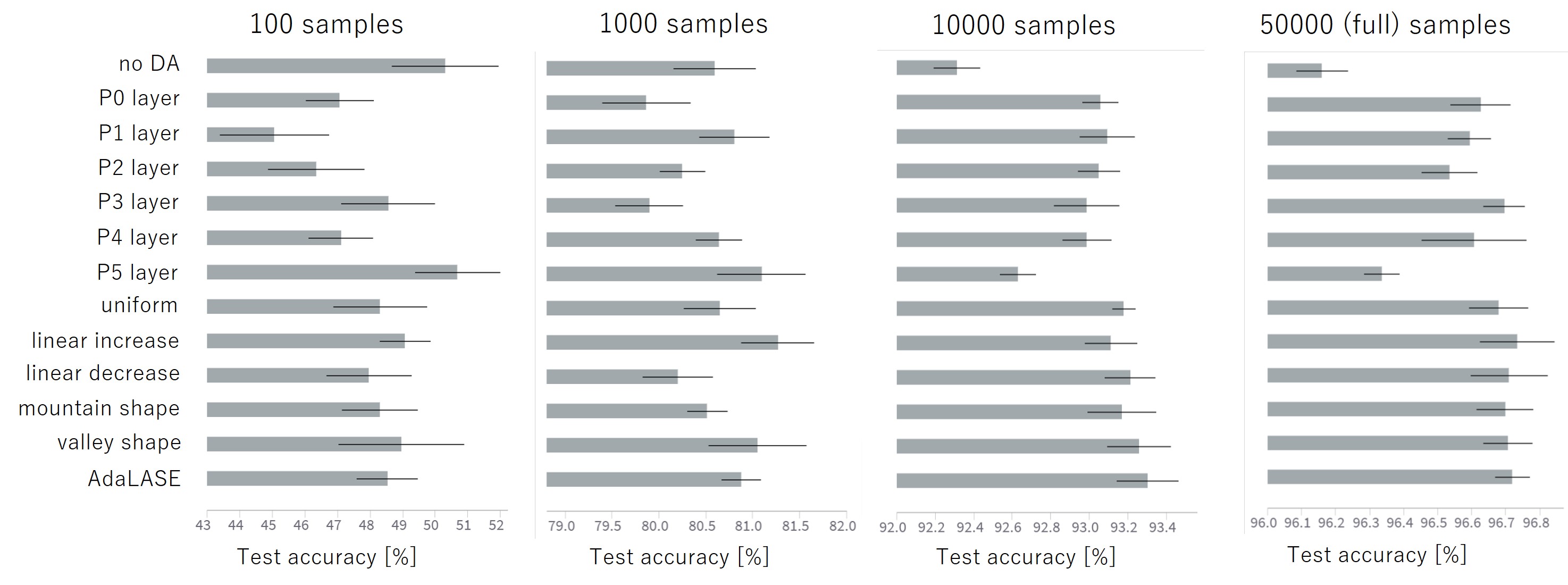}
}\\
\subfigure[Translation]{
\includegraphics[width=4.3in]{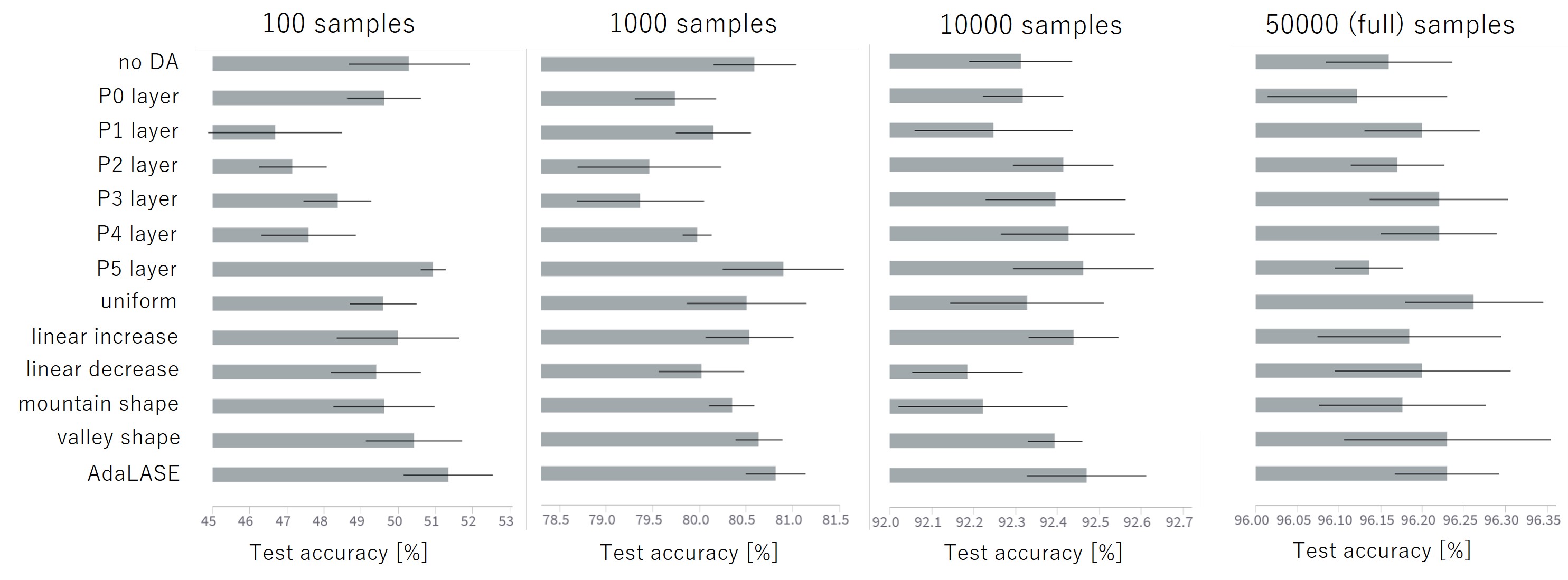}
}\\
\subfigure[CutMix]{
\includegraphics[width=4.3in]{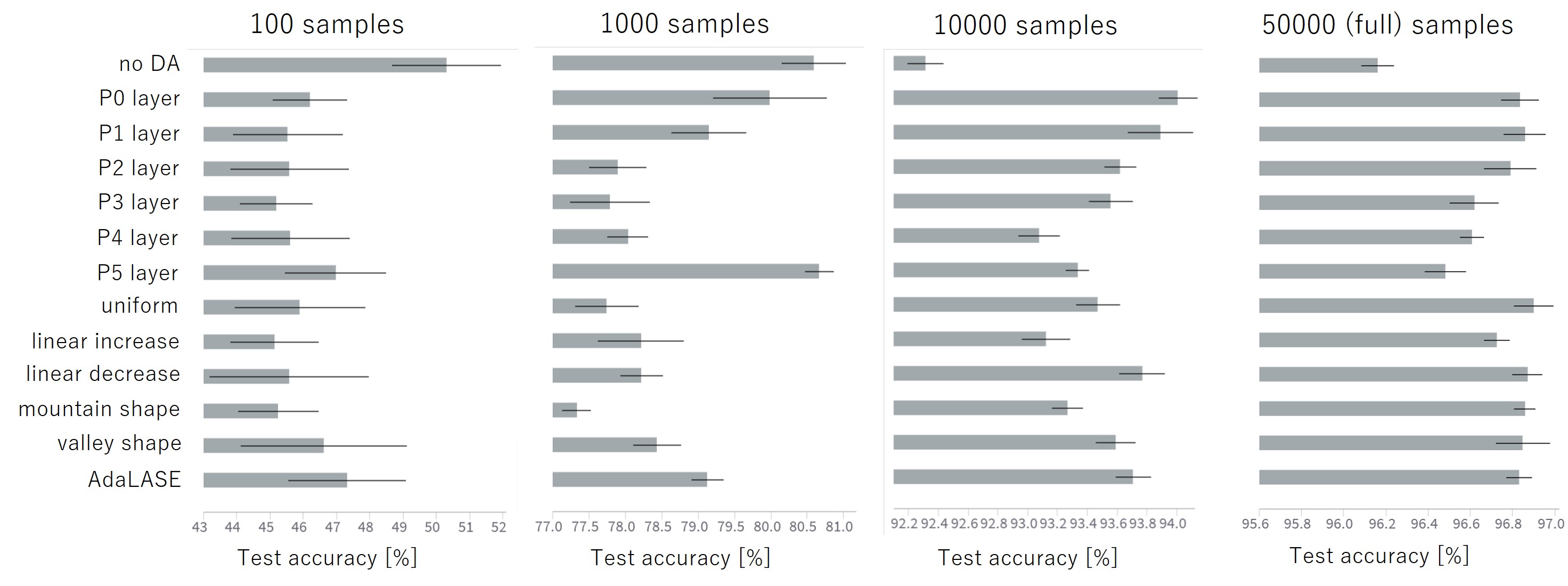}
}\\
\end{center}
\caption{Test accuracy relative to the number of training samples when training with DA in different layers. The pretrained ResNet18 model was trained on the CIFAR-10 dataset. The mean and standard deviation for five trials are shown.}
\label{fig:acc_num}
\end{figure}

Fig. 7 shows the test accuracy result obtained when DA was applied in various layers. P0 to P5 layers in the vertical axis denote the methods that fix the DA position to each layer during training. As can be seen, no DA method obtained high performance when the number of samples was small because models cannot learn fundamental features effectively if deformation is applied to a small number of samples. As the number of samples increased, the performance of the uniform DA method improved, which suggests that increasing the diversity of DA is important in terms of achieving high generalization performance if the number of samples is sufficient. Note that there were suitable and unsuitable layers when the layers to apply data augmentation were fixed, and these layers differ for different DA methods or the number of samples. If DA is applied in unsuitable layers, e.g., the P3 and P4 layers for mixup, the performance degrades considerably. We found that applying DA in the optimal layers improved the test accuracy greatly compared to applying DA in the worst layers. The proposed AdaLASE method can avoid unsuitable layers and select suitable layers by changing the acceptance ratio; thus, it obtained high performance overall regardless of the DA methods or the number of samples.

\begin{table}[tb]
\caption{Best and worst layers for DA fixed to each layer depending on the sample size. These results were obtained from the same experiment shown in Fig. 7.}
\label{table:best_and_worst_layers}
\vskip 0.15in
\begin{center}
\begin{small}
\begin{sc}
\subtable[The best layer]{
\begin{tabular}{rcccc}
\toprule
 & {\scriptsize Mixup} & {\scriptsize Cutout} & {\scriptsize Translation} & {\scriptsize CutMix}\\
\hline
100 samples      & P5 & P5 & P5 & P5\\
1000 samples     & P5 & P5 & P5 & P5\\
5000 samples     & P0 & P4 & P4 & P0\\
10000 samples    & P0 & P1 & P1 & P0\\
50000 samples    & P1 & P3 & P3 & P1\\
\bottomrule
\end{tabular}
}
\subtable[The worst layer]{
\begin{tabular}{rcccc}
\toprule
 & {\scriptsize Mixup} & {\scriptsize Cutout} & {\scriptsize Translation} & {\scriptsize CutMix}\\
\hline
100 samples      & P3 & P1 & P1 & P3\\
1000 samples     & P4 & P0 & P0 & P3\\
5000 samples     & P4 & P5 & P5 & P4\\
10000 samples    & P4 & P5 & P5 & P4\\
50000 samples    & P4 & P5 & P5 & P5\\
\bottomrule
\end{tabular}
}
\end{sc}
\end{small}
\end{center}
\vskip -0.1in
\end{table}

Table 2 shows the best and worst layers when the DA position was fixed to each layer. Note that these results were obtained from the experiments illustrated in Fig. 7. For the six positions in ResNet18 shown in Fig. 1(b), the layers that yielded the highest and lowest test accuracies are shown for each DA method and each number of training samples. From these results, we can observe different trends in layer suitability in terms of the number of samples, although trends are less observable in terms of the DA methods. For a small number of samples (P5), i.e., the layer that is close to the output layer, was the best layer. In contrast, when the number of samples was large, this layer was actually the worst layer. This difference can be interpreted as follows: DA in layers that are close to the input layer largely affects weights by feedforward propagation and changes to features acquired by training. This tends to deform the features inappropriately and causes underfitting when the number of samples is small. This effect may have become stronger due to the transfer of learning because the features acquired by pretraining can be collapsed. In contrast, for a large number of samples, DA should be applied in layers that are close to the input layer in order to modify the features to adapt to new and diverse training data.

\section{Conclusion}
Through a comprehensive set of experiments using different experimental settings, we found that suitable layers for DA differ depending on the experimental configuration. In particular, the major trend observed in this study is that effective layers for DA are strongly dependent on the number of training samples. When the number of samples was small, e.g., 100 samples, DA applied in unsuitable layers reduced the accuracy, and training without DA yielded high performance in most cases. In contrast, as the number of samples increased, DA applied in suitable layers became more significant, and the uniform DA method exhibited good performance. Applying DA in a suitable layer provided higher test accuracy than applying DA in an unsuitable layer, and identifying trends in suitable layers makes hyperparameter search for DA layers easier.

To search for suitable layers automatically, we proposed the AdaLASE method, which increases the acceptance ratio for suitable layers using gradient descent and is designed with a dynamic algorithm to reduce computational costs. The proposed AdaLASE method obtained test accuracy results that were close to the highest accuracy in many cases; thus, the proposed method can reduce the computational costs of searching for suitable layers. In addition, by investigating the transition of the acceptance ratio during the training process, we verified that the proposed AdaLASE method tended to reduce the acceptance ratio for unsuitable layers, and the algorithm implemented in the proposed method functioned as expected.

Identifying the trends of suitable layers for each task and developing the proposed method to dynamically search these layers contributes to revealing the black box nature of deep learning methods and facilitates automated machine learning. In this study, we focused on layer suitability; thus, in the future, we plan to apply the proposed AdaLASE method to optimizing DA methods. In addition, multiple DA methods are frequently used simultaneously; thus, the proposed AdaLASE method should be extended to optimize multiple methods, e.g., simultaneous use of cutout, mixup, and translation. This will also be the focus of future work.

\section*{Acknowledgements}
This study is based on results obtained from projects JPNP20006 and JPNP14004, which were commissioned by the New Energy and Industrial Technology Development Organization (NEDO), and a project of JSPS KAKENHI Grant Number 23K16966.
In addition, RK acknowledges the funding support from JST FOREST (Grant Number: JPMJFR226Q).

\bibliography{latentDA}

\end{document}